\documentclass[10pt,twocolumn,letterpaper]{article}

\usepackage{wacv}
\usepackage{times}
\usepackage{epsfig}
\usepackage{graphicx}
\usepackage{amsmath}
\usepackage{amssymb}
\usepackage{comment}
\usepackage{here}

\newcommand{\fnote}[1]{}

\newcommand{\knote}[1]{{\color{red} \bf #1 \color{black}}}

\newcommand{\bnote}[1]{{\color{red} #1 \color{black}}}
\newcommand{\mnote}[1]{}
\newcommand{\onote}[1]{\color{blue} #1 \color{black}}
\newcommand{\snote}[1]{}
\newcommand{\kcut}[1]{}
\newcommand{\ocut}[1]{}
\newcommand{\scut}[1]{}
\newcommand{\jptext}[1]{}

\ifx\pdfoutput\undefined
\else
 \usepackage{CJKutf8}
 \renewcommand{\fnote}[1]{}
 \renewcommand{\knote}[1]{}
 \renewcommand{\mnote}[1]{}
 \renewcommand{\bnote}[1]{}
 \renewcommand{\onote}[1]{}
 \renewcommand{\jptext}[1]{}
\fi

\ifx\etal\undefined\newcommand{\etal}{{\it et al. }}\fi
\ifx\ie\undefined\newcommand{\ie}{{\it i.e.}}\fi
\ifx\eg\undefined\newcommand{\eg}{{\it e.g.}}\fi

\newcommand{\subsecref}[1]{Sec.~\ref{ssec:#1}}

\wacvfinalcopy 

\def\wacvPaperID{492} 

\ifwacvfinal\fi
\begin{document}

\title{CNN based dense underwater 3D scene reconstruction \\by transfer learning using bubble database}

\author{Kazuto Ichimaru$^{\dag}$ \,\,\, Ryo Furukawa$^{\ddag}$ \,\,\, Hiroshi Kawasaki$^{\dag}$\\
$^{\dag}$ Kyushu University, Fukuoka, Japan\\
$^{\ddag}$ Hiroshima City University, Hiroshima, Japan}

\maketitle
\ifwacvfinal\thispagestyle{empty}\fi

\begin{abstract}
Dense 3D shape acquisition of swimming human or live fish is an important 
    research topic for sports, biological science and so on.
For this purpose, active stereo sensor is usually used in the air, however it cannot be 
    applied to the underwater environment because of refraction, strong light 
    attenuation and severe interference of bubbles.
Passive stereo is a simple solution for capturing dynamic scenes at underwater environment, however 
the shape with textureless 
    surfaces or irregular reflections cannot be recovered.
Recently, the stereo camera pair with a pattern projector 
    for adding artificial textures on the 
    objects is proposed.
However, to use the system for underwater environment,
several problems should be compensated, 
\ie, disturbance 
    by fluctuation and bubbles. 
Simple solution is to use
convolutional neural network for stereo to cancel 
    the effects of bubbles and/or 
    water fluctuation. Since it is not easy to train CNN with small size of 
    database with large variation, we develop a special bubble 
    generation device to efficiently create real bubble database of multiple 
    size and density.
In addition, we propose a transfer learning technique for multi-scale CNN to 
    effectively remove bubbles and  projected-patterns on the object.
Further, we develop a real system and actually captured live swimming human, 
    which has not been done before.
Experiments are conducted to show the effectiveness of our method compared 
    with the state of the art techniques.
%
%

\end{abstract}

\section{Introduction}
\label{sec:intro}

Dense 3D shape acquisition of swimming human or live fish is an important 
    research topic for sports, biological science and so on.
Passive stereo is a common solution for capturing 3D shapes because 
of its advantage on simplicity; \ie, it only requires two cameras.
In addition, since the shapes are recovered only from a pair of stereo images, it can 
capture moving or deforming objects.
One severe problem on passive stereo is instability, \ie, it fails to capture 
objects with textureless surfaces or irregular reflection.
To overcome the problem, using a pattern projector to add an artificial texture 
onto the objects has been proposed.
At underwater environments, there are additional problems for shape 
reconstruction by the system, such as 
refraction and disturbance by fluctuation and bubbles.
Further, since original textures of objects are interfered by bubbles and projected 
    patterns, they should be removed for obtaining
    original texture. 
For refraction issue, a depth-dependent calibration where refractions are 
approximated by a lens distortion of a center projection model is proposed~\cite{Kawasaki:WACV17}.
For disturbance issue, recently a convolutional neural network (CNN)-based stereo is proposed~\cite{Ichimaru:3DV2018}. 
However, those previous techniques still have some holes in their 
 results when 
bubble size is large, because the shapes are irregular and partially transparent. In addition, all the experiments are 
only conducted with a small water tank under controlled lighting condition. 

In this paper, we propose a transfer learning based CNN stereo as well as 
efficient construction of bubble database for the purpose; we develop a special bubble generation device to create a bubble database containing multiple size and density of bubbles.
For the texture recovery, we also propose a CNN-based method for projected-pattern removal and bubble-canceling method.
Since it is great labor to prepare task-specific dataset in extreme environment, 
we develop an unsupervised learning approach for texture recovery.
Further, we develop a real system to
capture live swimming human in a pool where lighting and 
other conditions are unknown and cannot be controlled.

Experimental results are shown to prove the effectiveness of our method by 
comparing the results with the previous 
methods~\cite{chang2018pyramid,Ichimaru:3DV2018,mccnn}. 
We also conduct demonstration to show the reconstructed sequence of swimming human.
Main contributions of the proposed technique are as follows:

\begin{enumerate}
  \setlength{\parskip}{0cm} 
  \setlength{\itemsep}{0cm} 
\item A multi-scale CNN-based stereo with transfer learning technique 
 specialized for underwater environment is proposed.
\item An unsupervised multi-scale CNN-based bubble and projected pattern removal method 
		  specialized for underwater environment is proposed
\item A special bubble generation device to create original database containing wide variety of bubble for transfer 
		  learning are developed; the database is plan to be public available.
\item The proposed technique is applied to live swimming human to recover the 
		  dynamic 3D shapes to confirm 
		  feasibility and practicability of the method.
\end{enumerate}

\section{Related work}
\label{sec:related}

For refraction problem, there are generally two types of solutions;
one is geometric approach and the other is approximation-based approach.
Geometric approach is based on physical models such as refractive index, distance to refraction interface, and normal of the interface
~\cite{Agrawal:CVPR2012,jordt2013refractive,kawahara:ICCV2013}.
Those techniques can calculate genuine light rays if parameters are correctly estimated and 
interface is completely planar, however, they are usually impractical. Further, 
the non-central projection camera model is not suitable for shape 
reconstruction in theory.
On the other hand, approximation approach converts captured images into central 
projection images by lens distortion and focal length adjustment~\cite{ferreira2005stereo}.
They assumed focal point moved backward to adjust light paths to be as linear as possible, then remaining error was treated as lens distortion.
It works well in most cases, but in specific case it fails because refractive 
distortion depends on depth and effective range of depth is not thoroughly
analyzed yet.

\begin{figure}[t]

\begin{minipage}[b]{0.5\linewidth}
  \centering
  \centerline{\includegraphics[width=4.0cm]{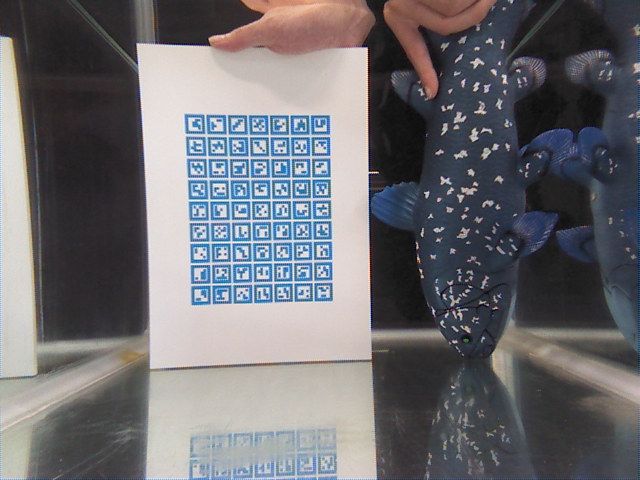}}
\end{minipage}
\hfill
\begin{minipage}[b]{0.49\linewidth}
  \centering
  \centerline{\includegraphics[width=4.0cm]{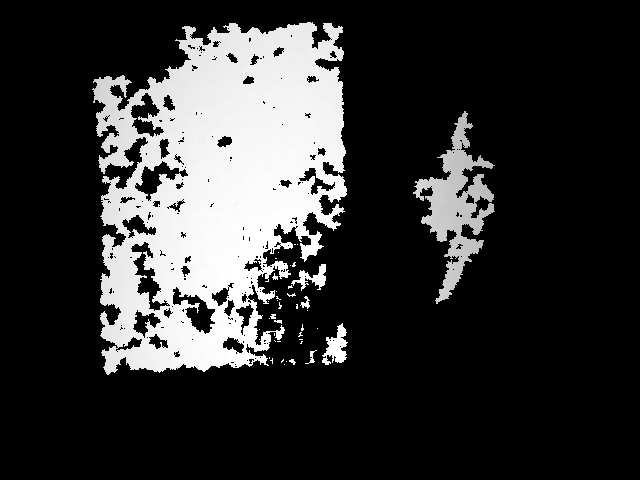}}
\end{minipage}

\caption{Captured RGB image and depth image by Kinect v1. Distance to the targets is 70cm.}
\label{fig:kinect}
\vspace{-0.0cm}
\end{figure}

In terms of light attenuation and disturbance problem for water medium,
light transport analysis has been conducted~\cite{Kutulakos:PAMI16,mukaigawa2010analysis}.
Narasimhan \etal proposed a structured-light-based 3D scanning method for 
strong scattering and absorption media based on light transport analysis~\cite{narasimhan2005structured}.
For weak scattering media, Bleier and N\"uchter used cross laser projector which only achieved sparse reconstruction~\cite{bleier}.
To increase density, Campos and Codina projected parallel lines with DOE to capture underwater objects with one-shot scan~\cite{massot2014underwater}.
Kawasaki et al. proposed a grid pattern to capture more dense 
shape with one-shot scan~\cite{Kawasaki:WACV17}.
One drawback of those one-shot scanning techniques is that 
reconstruction tends to be unstable even if light attenuation and disturbances 
are not so strong because sensitivity of pattern detection is high for 
subtle change of projected pattern.
Some research such as \cite{Anwer:Access2017} used infrared structured light or ToF sensors, 
but infrared attenuates rapidly in water as shown in Fig. \ref{fig:kinect}, and is not practical.

One simple solution is to apply passive stereo which is not much affected by those effects.
In the air, to increase the stability, Konolige investigated how to add active pattern to 
the passive stereo system~\cite{konolige} and there are also commercial products available~\cite{3DMD,realsense:sr300}. 
We focus its simplicity and stability to achieve dense dynamic reconstruction.

Recently, convolutional neural network (CNN) based stereo matching becomes popular.
\u{Z}bontar and LeCun proposed a CNN-based method to train network as a cost function of image patches~\cite{mccnn}.
Those techniques rather concentrate on textureless region recovery, but not 
noise compensation, which is a main problem for underwater stereo.
Since patch based technique is known to be slow, Luo \etal proposed a speeding-up technique by substituting FCN to inner product at final 
stage~\cite{Luo:CVPR2016}. Shaked and Wolf achieve high accuracy as well as fast 
calculation time by combining both FCN to inner product~\cite{Shaked:CVPR2017}.
To fundamentally solve the calculation time, end-to-end approach called DispNet is proposed, but accuracy is not so high~\cite{Mayer:CVPR2016}.
Another problem for patch based CNN stereo is that it is severely affected by 
obstacles, image degradation or various scaling.
Recently, multi-scale CNN 
technique is proposed to solve the patch size problem. Nah \etal proposed a 
method for debluring~\cite{Nah:CVPR2017}, Zhaowei \etal proposed a method for 
dehaze~\cite{Cai2016AUM} and Li \etal proposed a method for object recognition~\cite{li2017reside}, and
Yadati \etal, Lu \etal, Chen \etal, and Ye \etal~\cite{PramodYadati:ICMVA2017,HaihuaLu2018,JiahuiChen:ICIP2016,Ye:Access2017} used multi-scale 
features for CNN-based stereo matching. 
Chang and Chen extended multi-scale CNN stereo to end-to-end network called PSMNet~\cite{chang2018pyramid}, and achieved higher accuracy, 
but handleable resolution is very limited because of huge memory consumption.
Ichimaru \etal used FCN after multi-scale feature extraction to wisely integrate them~\cite{Ichimaru:3DV2018}, but they used only general stereo dataset.

Collection of huge data for learning is another open problem for CNN-based stereo 
techniques. For solution, Zhou \etal proposed a technique without using ground 
truth depth data, but LR consistency as a loss function~\cite{Zhou:ICCV2017}. Tonioni \etal 
proposed a unsupervised method by using existing stereo technique as an 
instruction~\cite{Tonioni:ICCV2017}.
Tulyakov and Ivanov proposed a multi-instance learning (MIL) method by using 
several constraints and cost functions~\cite{Tulyakov:ICCV2017}.
However in general, unsupervised learning is instable compared to supervised learning.
DispNet~\cite{Mayer:CVPR2016} and PSMNet~\cite{chang2018pyramid} are trained with generated images based on computer graphics,
but transfer learning with natural images is necessary since computer graphics is not realistic enough to learn noises or camera characteristics.
In this research, we created original stereo dataset and a special device for data augmentation 
which reproduces underwater environment for transfer learning.

CNNs are also popular in the field of image restoration and segmentation.
In underwater environment, there are several noises, such as bubbles or shadows 
of water surfaces. In addition, projected pattern onto the target object is also 
a severe noise.
To remove such a large noise,  inpainting method based GAN is 
promising~\cite{Iizuka:SIGGRAPH16,ChenyuYou2018}. However, since resolution of generative approach is 
basically low, noise removal approach is better fit to our purpose. For efficient 
noise removal, shallow CNN based approach using residual is proposed~\cite{He:CVPR2016}. 
Liu and Fang propose end-to-end architecture using WIN5RB network~\cite{Peng:arXiv2017} which outperform others. 
We also use this technique, but data collection and multi-scale extension is our original.
Further, we also propose unsupervised learning approach to overcome the difficulties preparing such task-specific dataset.

\section{System and algorithm overview}
\label{sec:overview}

\subsection{System Configuration}
\label{ssec:sysconf}
\begin{figure}[t]

\begin{minipage}[b]{0.5\linewidth}
  \centering
  \centerline{\includegraphics[width=4.3cm]{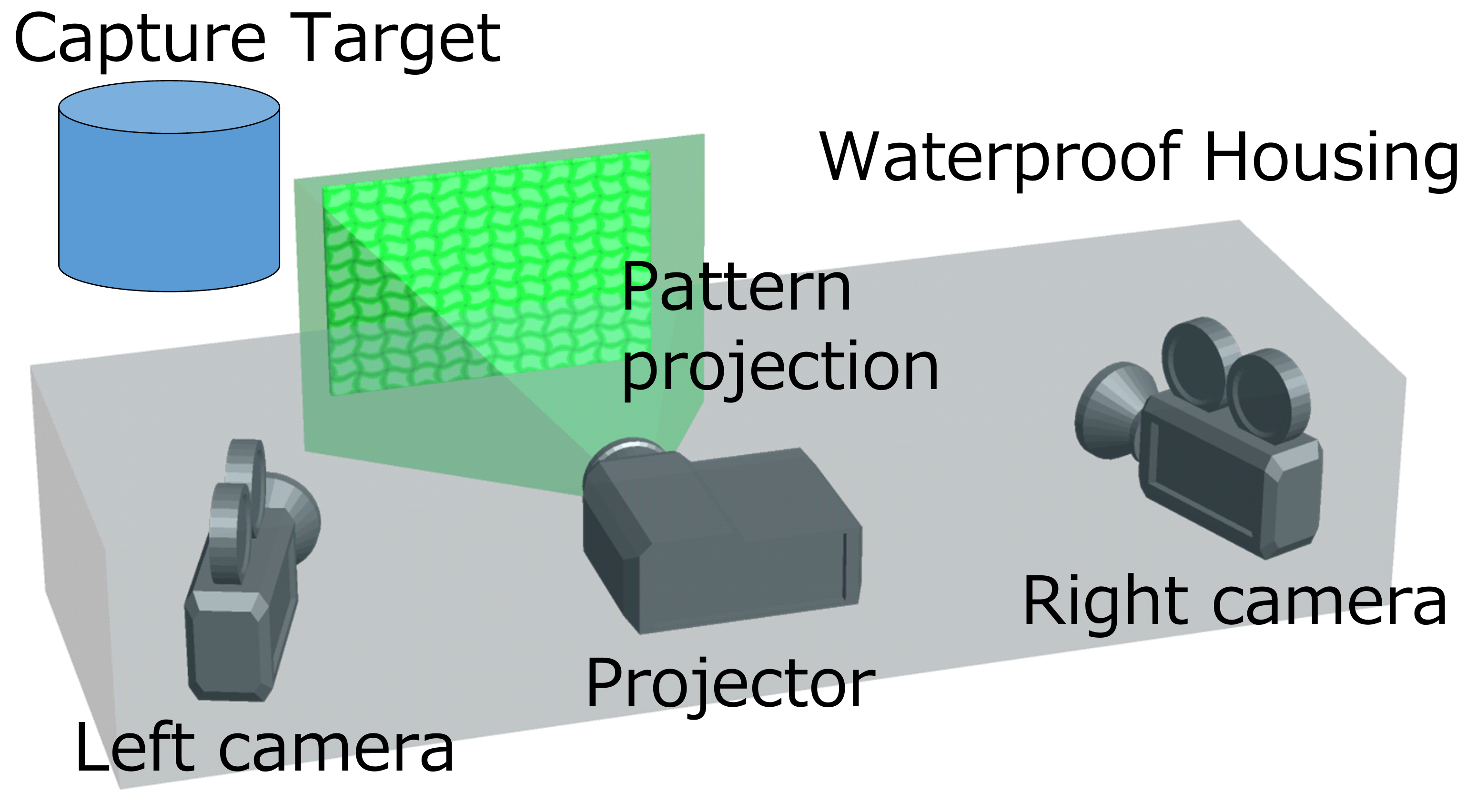}}
\end{minipage}
\hfill
\begin{minipage}[b]{0.49\linewidth}
  \centering
  \centerline{\includegraphics[width=3.5cm]{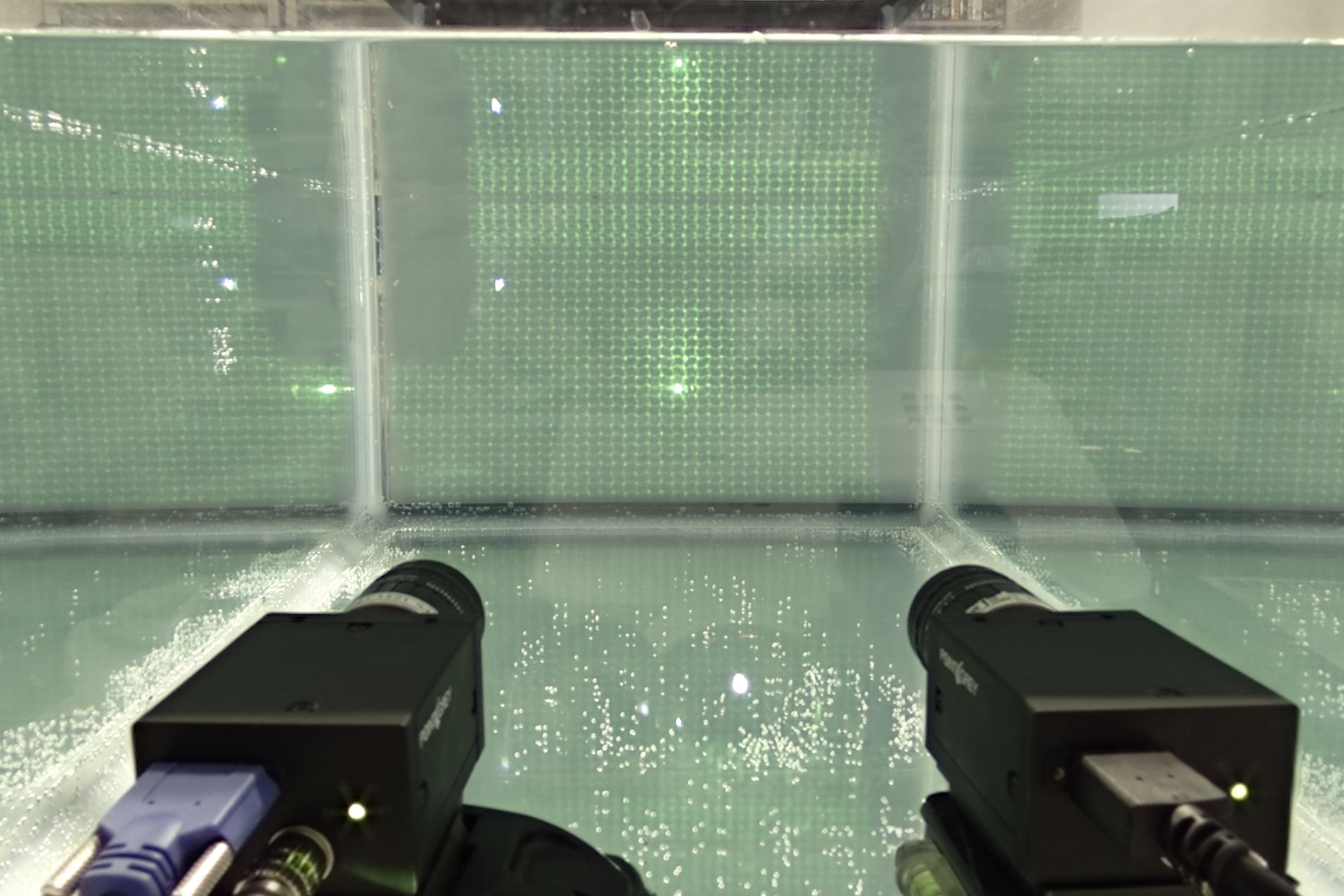}}
\end{minipage}

\caption{{\bf Left: }Minimum system configuration of the proposed algorithm. 
{\bf Right: }Our experimental system for evaluation where two cameras and a projector are set outside a water tank.}
\label{fig:sysconf}
\vspace{-0.0cm}
\end{figure}

The proposed system includes
two cameras and one projector as shown in Fig.~\ref{fig:sysconf}.
We prepare two systems for experiments.
One is for evaluation purpose where two cameras and a projector are set outside a water tank.
The other is a practical system where devices are installed into a waterproof 
housing. For the both systems, the optical axes of the cameras are set orthogonal to housing surface so 
that error by refraction approximation is minimized. 
The two cameras are synchronized to capture dynamic scenes.
In terms of the pattern projector to add textures onto the objects,
no synchronization is required
since the pattern is static. 

\subsection{Algorithm}
\label{ssec:algo}

\begin{figure}[t]
\begin{minipage}[b]{1.0\linewidth}
  \centering
  \centerline{\includegraphics[width=8.0cm]{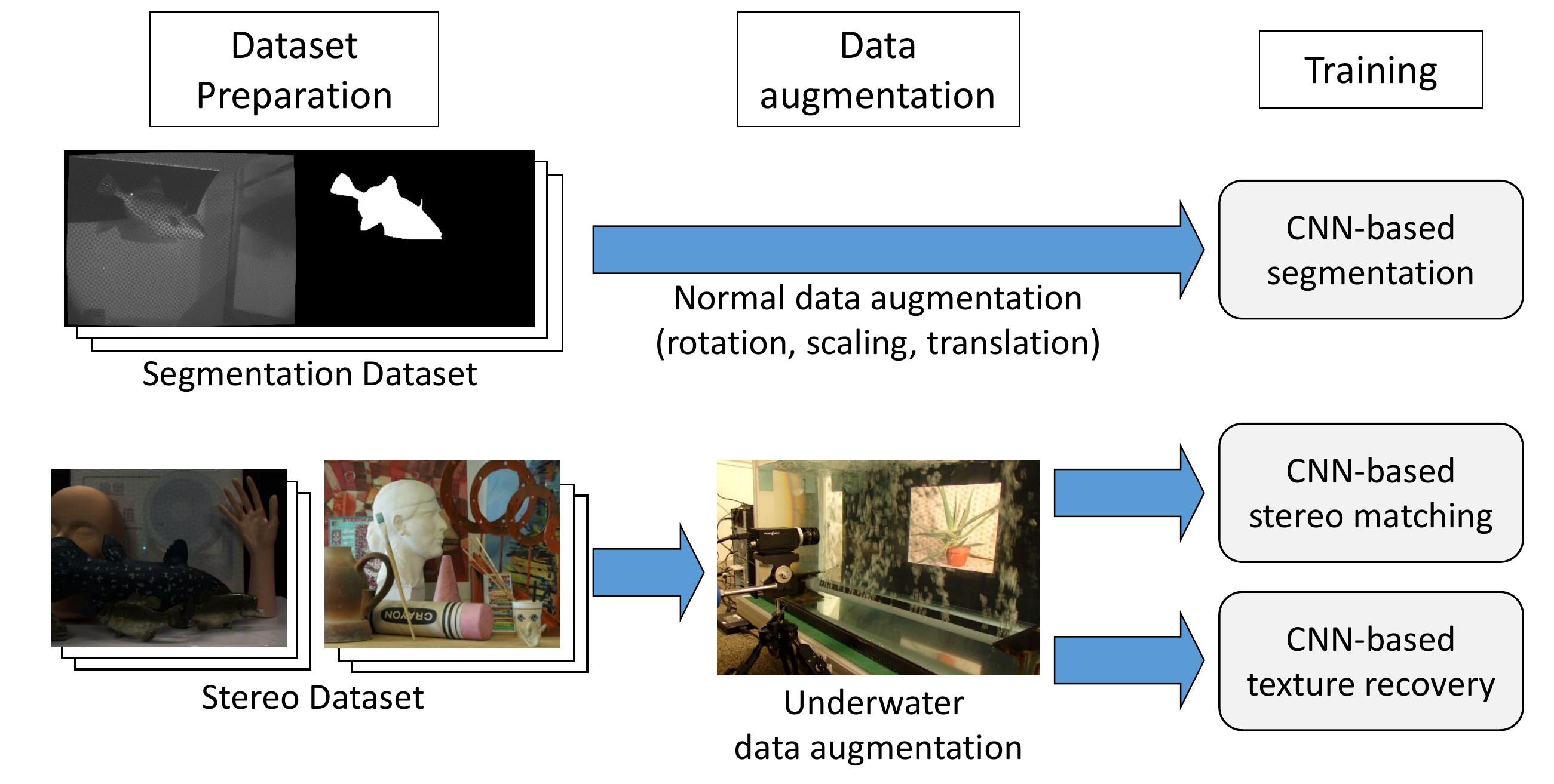}}
\end{minipage}
\hfill
\begin{minipage}[b]{1.0\linewidth}
  \centering
  \centerline{\includegraphics[width=8.0cm]{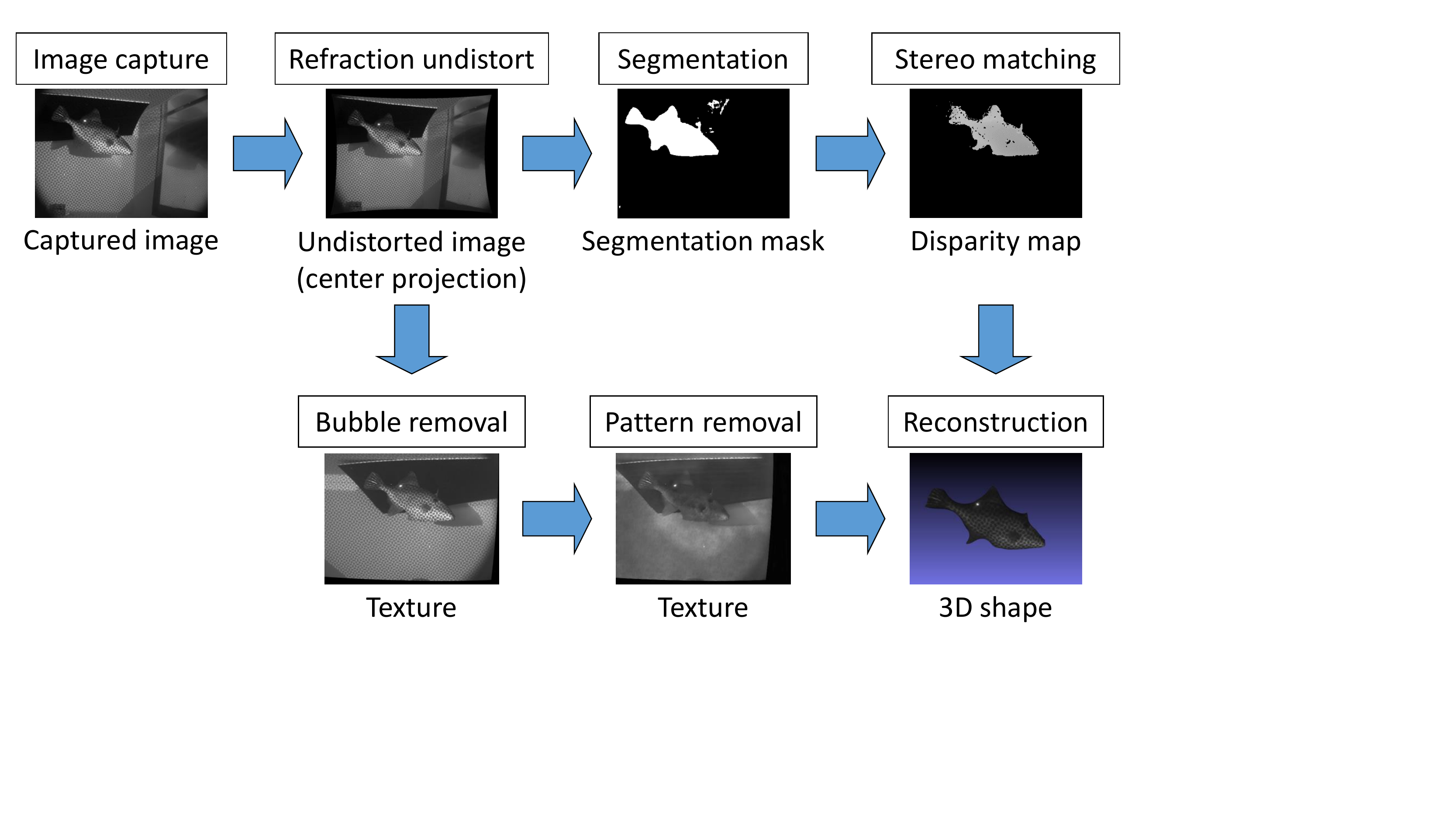}}
\end{minipage}

\caption{ Overview of the algorithm. }
\label{fig:overview}
\vspace{-0.0cm}
\end{figure}

The flow of our algorithm is shown in Fig.~\ref{fig:overview}.
In learning phase, several CNNs such as CNN-based segmentation, CNN-based stereo matching, and CNN-based texture recovery
are trained for robustness against underwater disturbances as shown in 
Fig.~\ref{fig:overview} (top).
First, CNN-based segmentation network is trained to detect reconstruction target region.
It can be trained with large image dataset, or small dataset created from captured images in reconstruction phase. In the method, we manually created the mask data for learning.
For CNN Stereo, proper stereo dataset suitable for assuming application is 
prepared (\eg, fish and human images, in our case) without bubble.
We also create special dataset which reproduces underwater environment by using a 
special bubble generation device for transfer learning purpose.
CNN-based stereo matching network is efficiently trained by using both datasets.
CNN-based texture recovery networks are also trained with prepared dataset.

Reconstruction process is shown in Fig.~\ref{fig:overview} (bottom).
First, the camera pair is calibrated. 
The refractions in the captured images are modeled and canceled by 
center projection approximation by depth-dependent calibration~\cite{Kawasaki:WACV17}.
In the measurement process, the targets are captured with stereo cameras.
Pattern illumination is projected onto the scene for adding features on it. 
From captured images, target regions are detected by a CNN-based segmentation technique, where only target regions are extracted.
Then, a stereo-matching method is applied to the target regions. 
In our technique, a CNN-based stereo is applied to increase stability under the 
condition of dimmed patterns, disturbances by bubbles, and flickering shadows. 
Then, 3D points are reconstructed from disparity maps estimated by the stereo algorithm.
Outliers are removed from the point cloud and 
meshes are recovered by Poisson equation method~\cite{Kazhdan:EGSGP06}.
Since textures are degraded by bubbles and projected patterns, they are 
efficiently recovered 
by CNN-based bubble canceling and pattern removal technique.
Using the recovered 3D shapes and textures, 
we can render the dynamic and textured 3D scene. 

\section{Stereo reconstruction using CNNs}
\label{sec:cnnstereo}

To deal with bubbles and water fluctuation disturbing the images, 
we create special database containing multiple size and density of bubbles at 
real underwater environment for transfer learning (\subsecref{dataset}).
By using the data, we apply CNN-based target region extraction technique (\subsecref{segmentation})
and multi-scale CNN stereo (\subsecref{msc}).

\subsection{Underwater stereo dataset created by special bubble generation device}
\label{ssec:dataset}

\begin{figure}[t]

\begin{minipage}[b]{0.49\linewidth}
  \centering
  \centerline{\includegraphics[width=4.0cm]{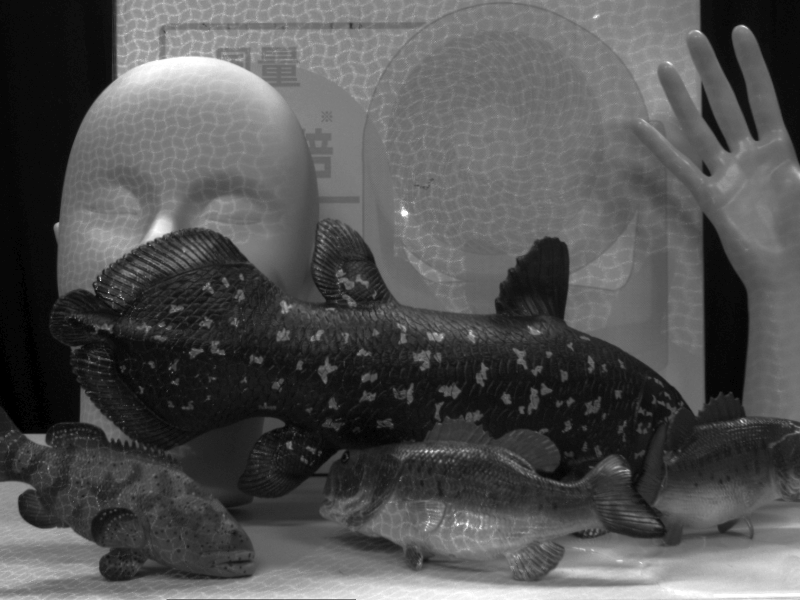}}
\end{minipage}
\hfill
\begin{minipage}[b]{0.49\linewidth}
  \centering
  \centerline{\includegraphics[width=4.0cm]{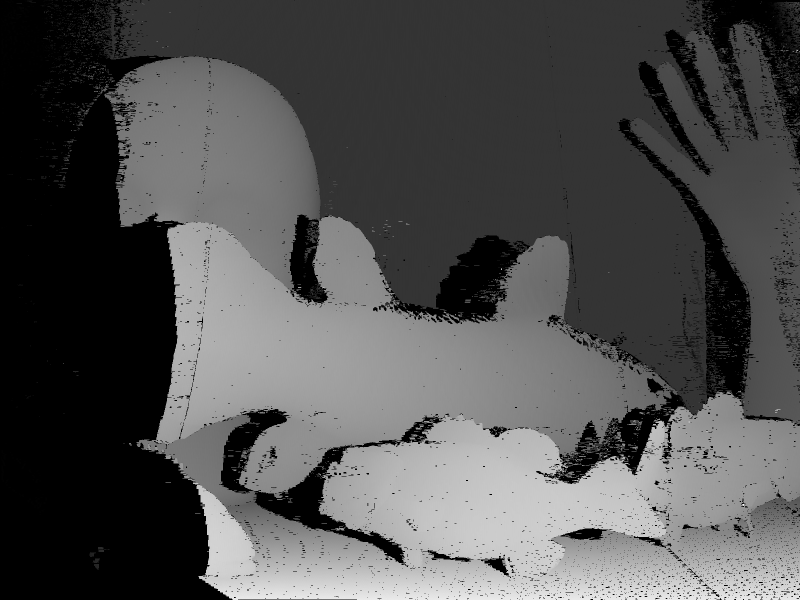}}
\end{minipage}

\caption{ {\bf Left:} An example of images included in Coel Dataset. 
{\bf Right:} An example of disparity map. }
\label{fig:dataset}
\vspace{-0.0cm}
\end{figure}

First, we create basic training datasets for stereo with projected pattern as follows.
Since we assume measurement of swimming human and fish as our purpose,
dataset which includes fish-like and human skin-like objects is necessary.
Thus, we prepared model of coelacanth, largemouth bass, goliath grouper, mannequin head and hand.
We prepared two cameras and one projector, and captured the above models and some additional objects 
in the air with graycode techniques from two views.
Then we captured targets with two cameras changing room illumination and pattern projection condition.
We captured 8 target object set, 3 poses, 3 illumination, and 4 pattern projection condition, in total 288 stereo image pairs were acquired.
We also created ground truth disparity map from captured graycode images as shown in Fig.~\ref{fig:dataset} (right).
The dataset is named ``Coel Dataset''.
An example of the image in the dataset is shown in Fig.~\ref{fig:dataset} (left).

\begin{figure}[t]

\begin{minipage}[b]{0.49\linewidth}
  \centering
  \centerline{\includegraphics[width=4.0cm]{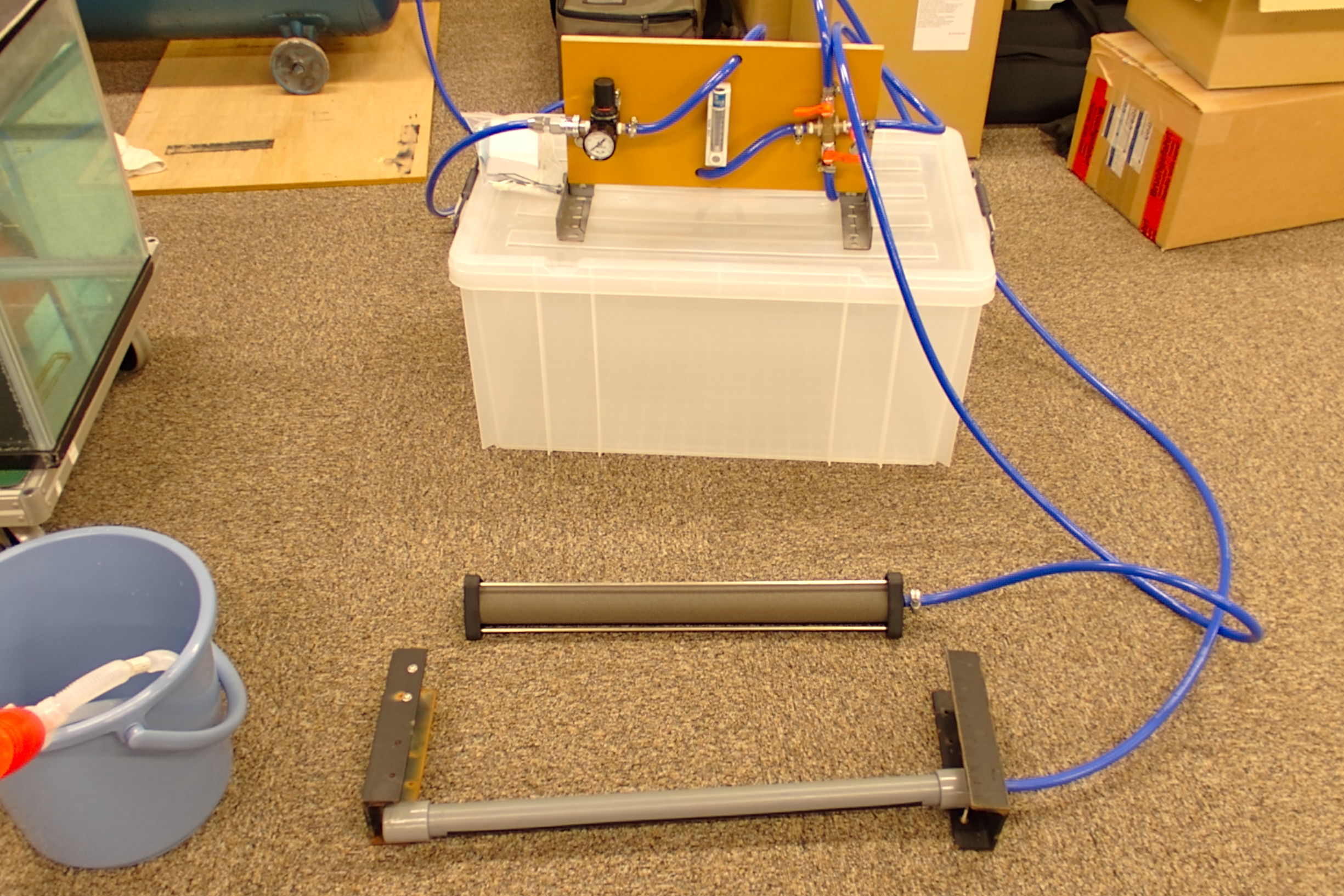}}
\end{minipage}
\hfill
\begin{minipage}[b]{0.49\linewidth}
  \centering
  \centerline{\includegraphics[width=4.0cm]{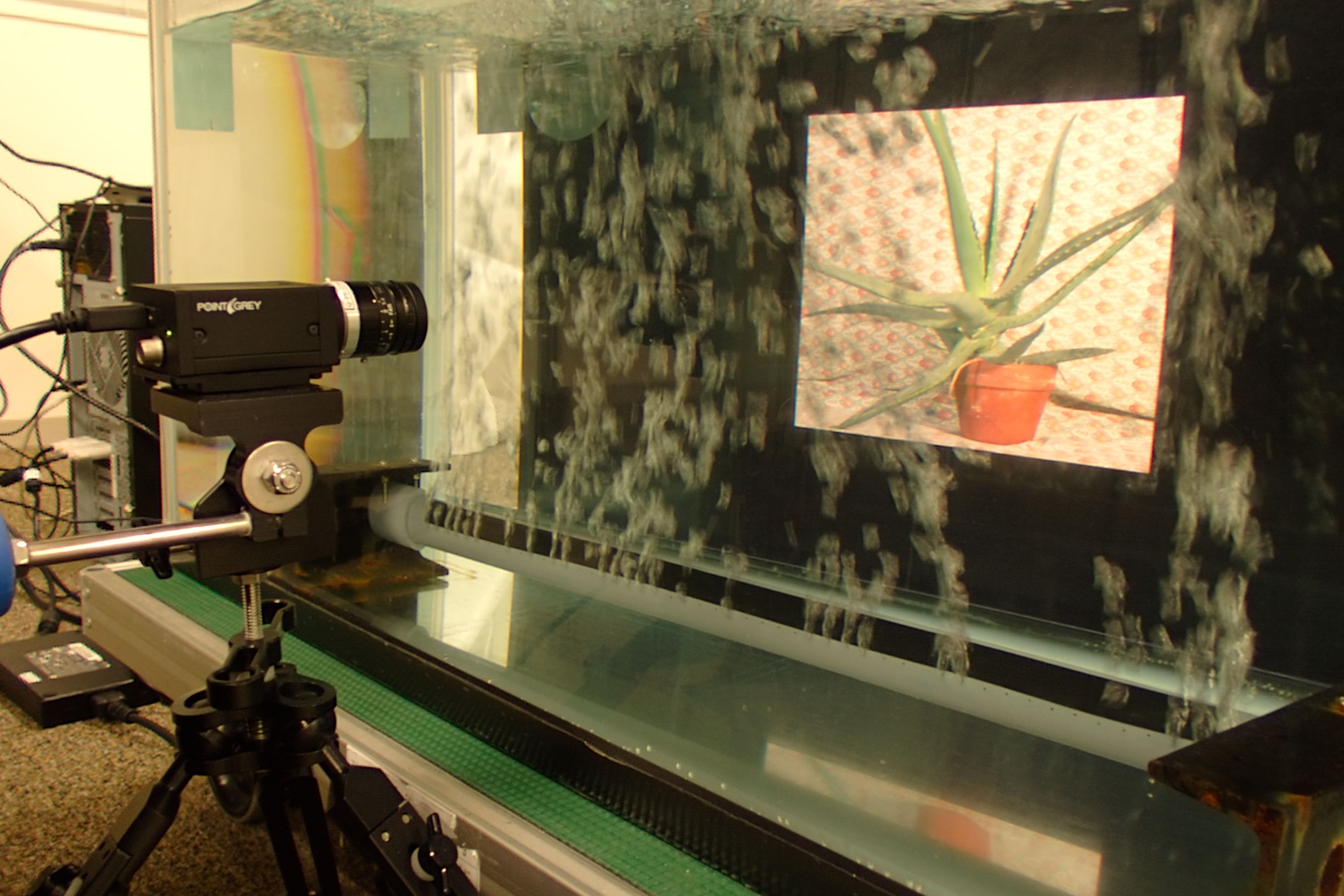}}
\end{minipage}

\caption{ {\bf Left:} Appearance of bubble generator.
{\bf Right:} Situation of bubble data augmentation. }
\label{fig:augmentation}
\vspace{-0.0cm}
\end{figure}

\begin{figure*}[t]

\begin{minipage}[b]{0.49\linewidth}
  \centering
  \centerline{\includegraphics[width=8.45cm]{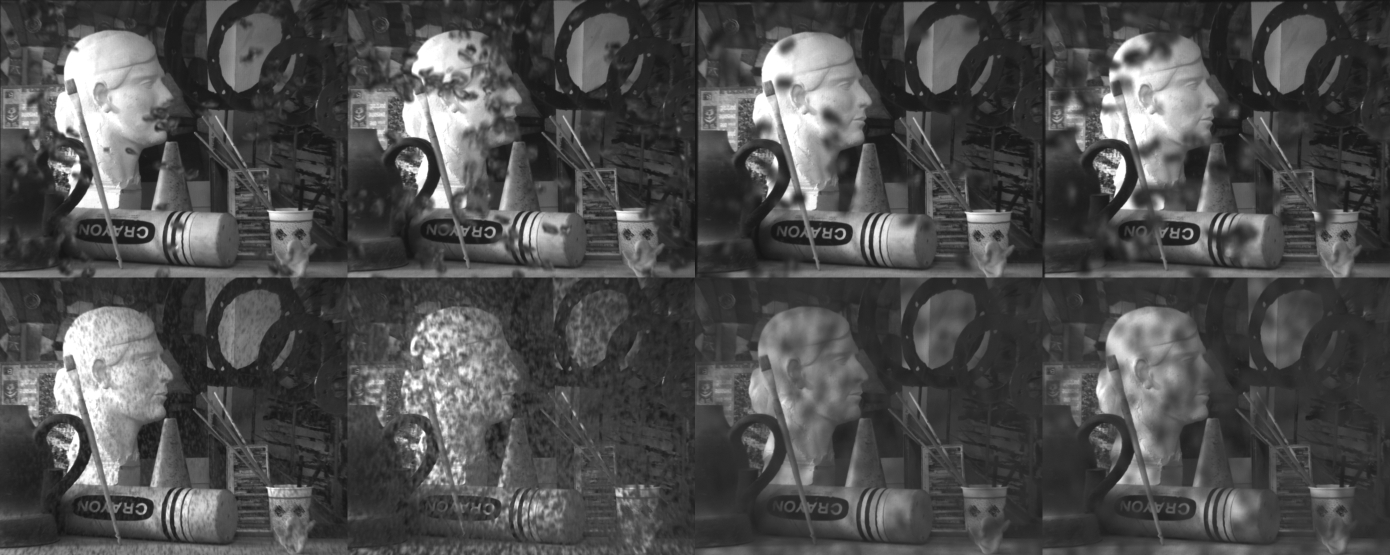}}
\end{minipage}
\hfill
\begin{minipage}[b]{0.49\linewidth}
  \centering
  \centerline{\includegraphics[width=9cm]{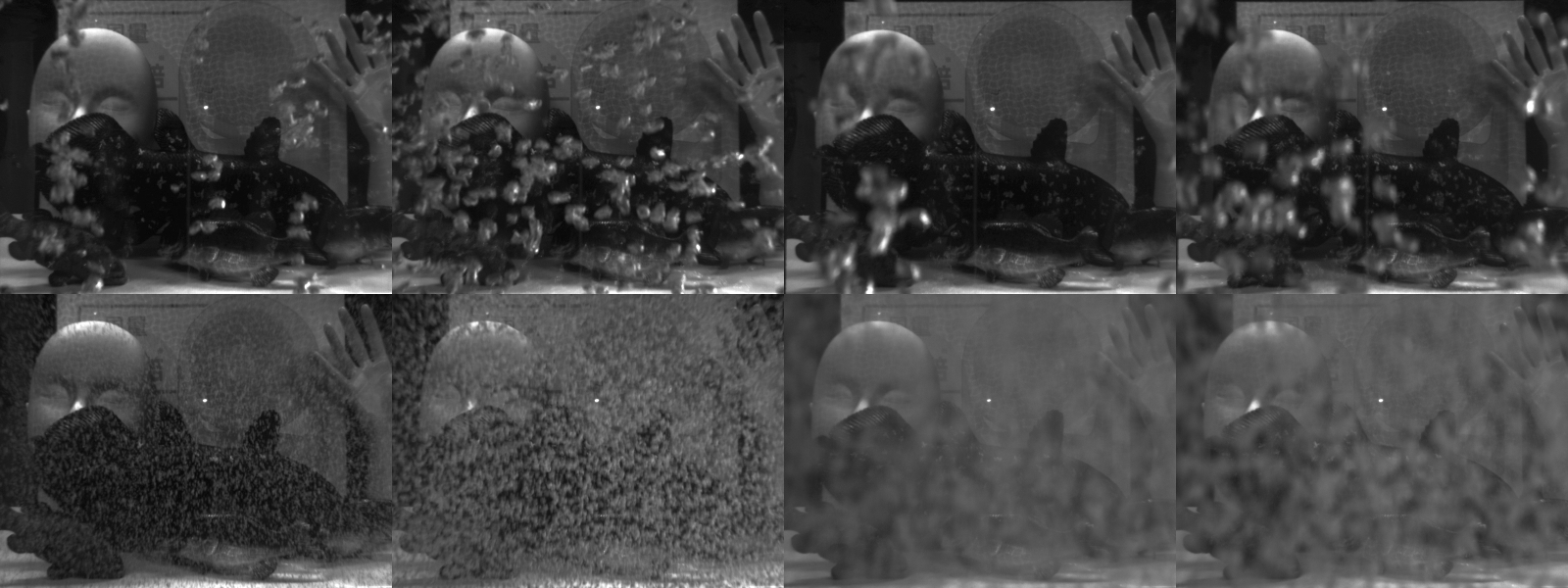}}
\end{minipage}

\caption{ Examples of augmented dataset with various types of bubbles. }
\label{fig:datasetexample}
\vspace{-0.0cm}
\end{figure*}

Then, we create special training datasets for underwater stereo as follows.
To effectively train the network with scene include bubbles, we develop a 
special bubble 
generator to reproduce underwater environment (Fig.~\ref{fig:augmentation}).
Since underwater bubbles have wide variety, it is necessary to generate various types of bubbles.
Our bubble generator can control bubble size, density, and generating position.
We used the device to augment stereo dataset with bubble.
Then, we placed a camera and a LCD monitor as they face each other, 
and placed water tank of $90\times45\times45$cm in-between them.
Water tank was filled with transparent water and bubble generator is submerged into the tank.
Graycode patterns were presented on the monitor and captured by the camera in order
to acquire 2D point correspondences between camera image plane and the monitor 
pixel position.
Then, arbitrary images of public available dataset were displayed on the monitor 
and captured by the camera while bubbles are generated in the tank.
We captured dataset in 2 bubble sizes, 2 generating positions and 2 densities, 
\ie, total 8 cases plus one no-bubble scene
as shown in Fig.~\ref{fig:datasetexample}.
Used images are Middlebury 2005, 2006 dataset and Coel Dataset which contains 918 images.
Note that such underwater stereo datasets including real bubbles do not yet 
exist and we will make the datasets public available; this is one of our 
important contribution of the paper.

\knote{
・注意
今回は単純に画像をマスクして入力するだけでなく, CUDAによるSGMにおいて探索範囲をマスク内に限定する制約を加えてある(ms-cnn-fcn).
End-to-Endは明示的にそのような処理を加えてはいないが, 多分真っ黒な領域は視差を出す必要が無いと学習してくれている
}
\subsection{CNN-based target-region extraction}
\label{ssec:segmentation}
For many applications, reconstruction targets are recognizable, such as swimming human in the water. 
In general, the wider the range of disparities considered in stereo-matching processes, 
the more possibilities exist, leading to wrong correspondences.
Thus, by extracting the target regions from the input images and 
reducing possibilities of matching outside the target regions, 
3D reconstruction process becomes more robust. 
In addition, 3D points of only required region can be obtained.

\begin{figure}[t]

\begin{minipage}[b]{0.24\linewidth}
  \centering
  \centerline{\includegraphics[width=2.0cm]{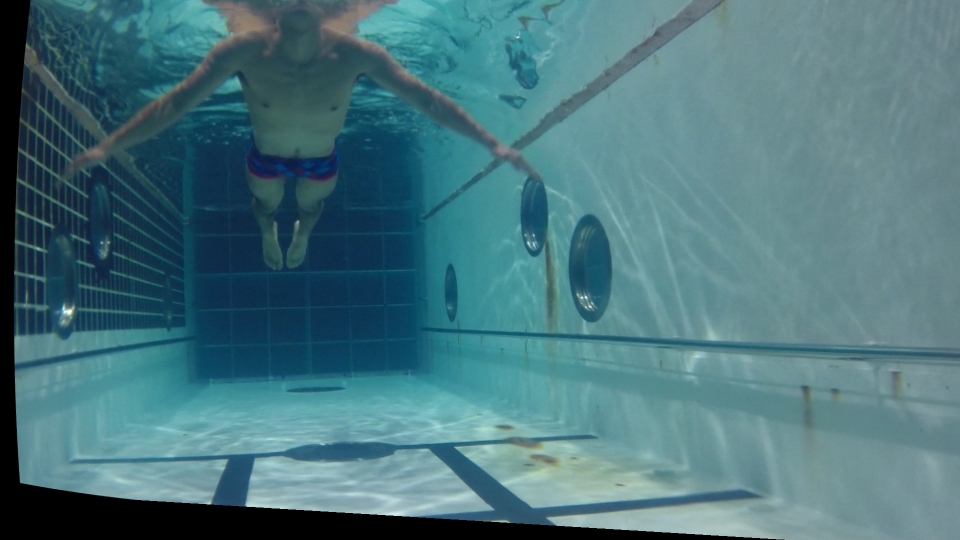}}
\end{minipage}
\hfill
\begin{minipage}[b]{0.24\linewidth}
  \centering
  \centerline{\includegraphics[width=2.0cm]{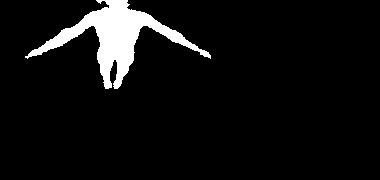}}
\end{minipage}
\hfill
\begin{minipage}[b]{0.24\linewidth}
  \centering
  \centerline{\includegraphics[width=2.0cm]{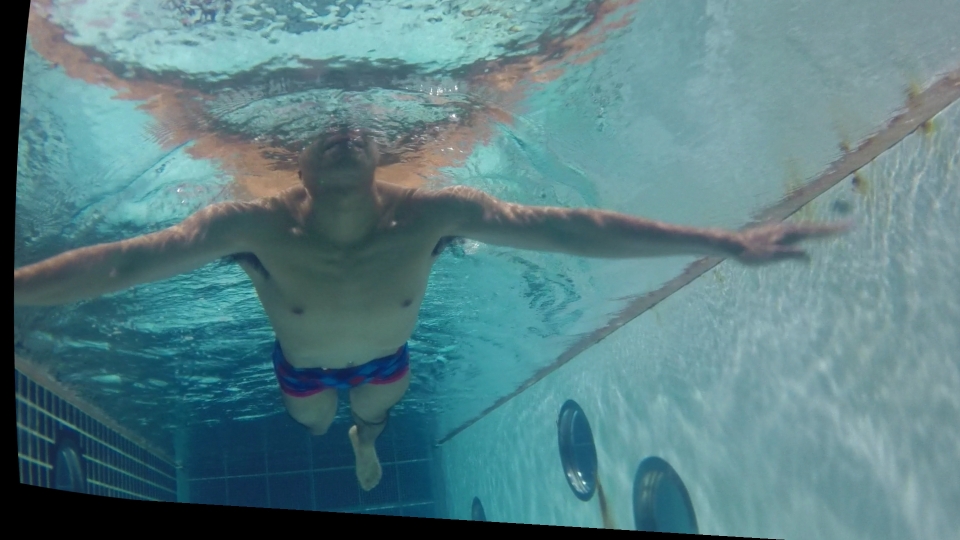}}
\end{minipage}
\hfill
\begin{minipage}[b]{0.24\linewidth}
  \centering
  \centerline{\includegraphics[width=2.0cm]{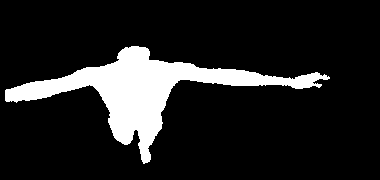}}
\end{minipage}

\caption{ An example of CNN segmentation results. }
\label{fig:seg}
\vspace{-0.0cm}
\end{figure}

To this purpose, 
we implemented an U-Net~\cite{ronneberger2015u}, an FCN with multi-scale feature extraction, and trained it for this task. 
%
We made training dataset from underwater image sequences.
From image sequences, 200 images were sampled and the 
target regions were masked with manual operations.
These training images were augmented by scalings, rotations, and translations.
As a result, we provide 2000 pairs of source images and target-region masks for training U-Net. 
We use cross entropy as loss function.
The trained U-Net is tested for large number of images,
 we obtained qualitatively successful results in most examples, even when camera was moving during capturing
 (Fig.~\ref{fig:seg}).

In the evaluation process, we have found that the numbers of resolution levels of the U-Net architecture is important.
By using only two or three levels of resolutions, we could not get sufficient results. 
We finally reached the conclusion that the U-Net with five levels of 
resolutions works effectively with our dataset.  

Using the obtained results, disparity search range can be limited and which improves robustness.
The search range limitation is implemented into CNN stereo matching described later, 
which takes mask images as input in addition to rectified images.

\subsection{Multi-scale CNN stereo for robustness against bubble}
\label{ssec:msc}

There are several peculiar phenomena in the water such as bubbles, shadows of water surface, 
refraction caused by difference of water temperature, and so on.
Those phenomena degrade the captured image and make bad effects on stereo matching.
Since those phenomena are explained by complicated process of both physics and optics, 
 it is difficult to 
solve analytically. In the paper, we adopt a 
learning based approach, \ie, CNN, for solution.

\begin{figure}[t]
	\begin{minipage}[b]{\linewidth}
		\centering
		\includegraphics[width=9.0cm]{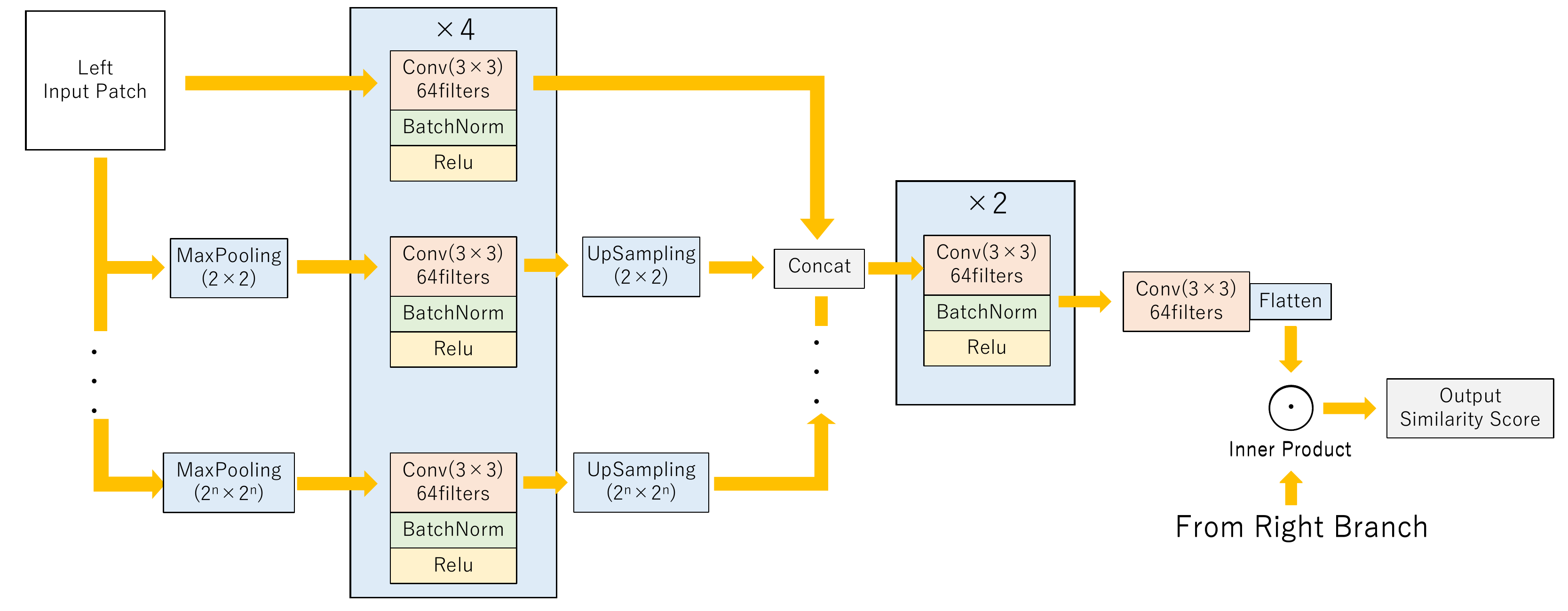}
	\end{minipage}
	
	\caption{ Network architecture of multi-scale CNN stereo. Left and right branches are symmetry. }
	\label{fig:network-stereo}
\vspace{-0.0cm}
\end{figure}

Since bubbles and water fluctuation have much larger structure than pixel scale,
basic CNN-based stereo matching technique does not work~\cite{mccnn}.
There are several papers which can handle a large spatial 
structure by multi-scale extension~\cite{PramodYadati:ICMVA2017,HaihuaLu2018,JiahuiChen:ICIP2016}
and they can recognize large scale information, 
however, such previous methods does not use enough number of scales, 
or utilize multi-scale information due to flattening of multi-scale features.
In the paper, we propose multi-scale CNN stereo architecture which takes multiple scale patches (more than three)
and process the retrieved features by Fully Convolutional Networks (FCN); this 
is not common for multi-scale CNN stereo.
The architecture is shown in Fig.~\ref{fig:network-stereo}.
It takes $44\times44$ (the size depends on the number of scales) patches from 
left and right rectified images for calculation.
In the matching process, first, patches are down-sampled by MaxPooling layers, and multiple scale patches are prepared.
Second, each patch is processed by former FCN to retrieve multi-scale features, respectively.
Then, retrieved multi-scale features are up-sampled to original scale, concatenated, 
and processed by FCN to integrate multi-scale information.
Finally, output tensors are vectorized and cosine distance between left and right vector is calculated to determine the similarity score.
Weights are shared between left and right branches to reduce memory consumption,
because multi-scale CNN architecture tend to consume large memory.
Cost volume is obtained by sweeping input images with patches.

Finally, post-processing are applied such as Semi-Global Matching 
(SGM)~\cite{SGBM} as well as 
 Left-Right consistency check to get final disparity map.
Post-processing is implemented by GPU and it achieved fast calculation as mentioned in experiment section.
Effectiveness of our architecture is confirmed in experiment section (\subsecref{robust}).


\section{Unsupervised texture recovery from bubble and projected pattern}
\label{sec:texture}

For real situations, the captured images are often severely degraded by underwater environments, such as bubble and other 
noises, as well as projected pattern on the object surface.
In order to remove such undesirable effects, we propose CNN-based solution.
In our technique, we focus on two major problems, such as bubbles and projected 
pattern. Although those two phenomena are totally different and have different 
optical attributes, it is common in the sense that appearances for both effects have a wide variation in scale.
Note that such wide variation depends on the distance between a target object, bubble and a projector.
Such a large variation of scale of them makes it difficult for removal by 
simple noise removal method.

\begin{figure}[t]
	\begin{minipage}[b]{\linewidth}
		\centering
		\includegraphics[width=9.0cm]{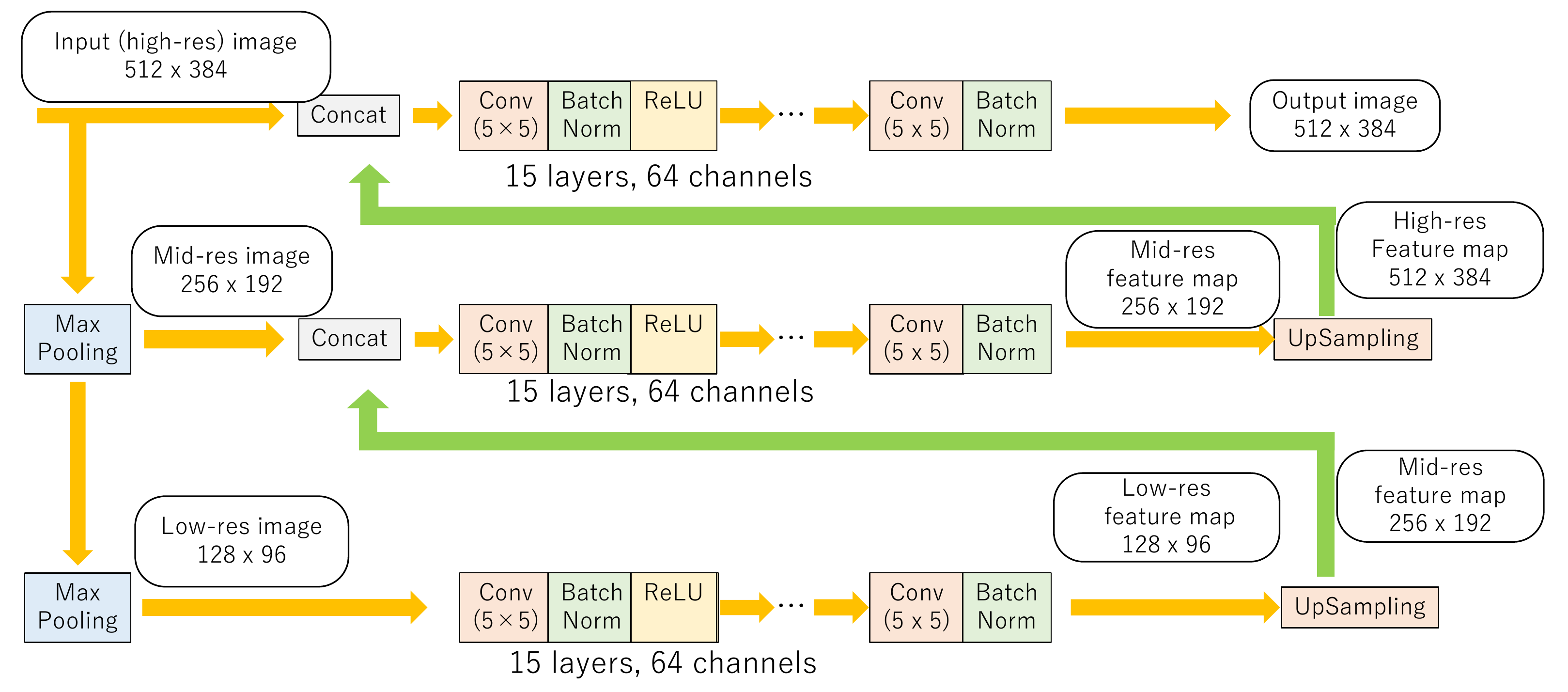}
	\end{minipage}
	
	\caption{ Network architecture of multi-scale CNN texture recovery. }
	\label{fig:network-texture}
\vspace{-0.0cm}
\end{figure}

Since multi-scale CNN is suitable to learn such a variation, we also use a 
multi-scale CNN for our bubble and pattern removal purpose.
The network for such obstacle removal is shown in Fig.~\ref{fig:network-texture}.
In the figure, it is shown that an original image is converted to three 
different resolutions and trained by independent CNN. Each output is up-sampled 
and concatenated to higher resolution.
This network is advantageous because it can handle a large structure of 
projected pattern, as well as it can be trained in a relatively short time.

For pattern removal training phase, we also used Coel Dataset as pattern removal training data which contains both images with and without pattern on same scene.
We used images with pattern as input, and images without pattern as ground truth.
Ground truth images were adjusted their brightness and contrast to fit to input images.
In addition, although we need to remove projected pattern on real fish, Coel Dataset does not contain such images.
Thus, we prepared raw fish (sea bream, filefish and chicken grunt) and submerged them into the water, 
then we captured them with and without pattern to make transfer dataset.
The pattern removal network was trained with the transfer dataset after trained with Coel Dataset.
In terms of bubble removal training, we used bubble augmented stereo dataset aforementioned in \subsecref{dataset} using no-bubble scene as ground truth.

Furthermore, we consider unsupervised learning approach to recovery correct texture even in case training dataset is insufficient.
To achieve it on pattern removal, first we trained pattern detection network to output the difference image between with and without pattern.
The network architecture is duplication of texture recovery network shown in Fig.~\ref{fig:network-texture}.
It can be trained in insufficient dataset situation because pattern detection is easier than pattern removal, 
or we can use alternative pattern detection method if necessary as \cite{Kawasaki:WACV17}.
Using pattern detection network, loss function is defined as below:
\begin{equation}
\begin{split}
MSE(in, R(in)) + \lambda \times MSE(D(R(in)), \vec{0}) \nonumber
\end{split}
\end{equation}
where $in$ means input image, $R$ and $D$ means output of pattern removal and detection respectively,
$MSE$ means mean squared error, and $\lambda$ is coefficient for balancing.
The loss function leads to pattern is not detected, \ie, pattern is removed, while keeping output image not far from input image.
Result of the function is back-propagated to only pattern removal network.
The network was also trained with Coel Dataset and fish transfer dataset.
Effectiveness of this unsupervised learning is confirmed in experiment section (\subsecref{texeval}).


\section{Experiments}

In order to evaluate our proposed method, we conducted several experiments.
First, robustness of Multi-scale CNN stereo against underwater disturbances is investigated in \subsecref{robust}.
Then, qualitative evaluation of texture recovery is conducted in \subsecref{texeval}.
Finally, we captured real swimming human sequence and reconstructed it by proposed method to confirm its feasibility in \subsecref{demo_human}.

\subsection{Evaluation of various CNN stereo techniques}
\label{ssec:robust}
We tested CNN-based stereo for underwater scene with bubbles.
For evaluation purpose, we prepared six implementations,
such as single-scale CNN stereo of \cite{mccnn} (mc-cnn), 
2-scale CNN stereo with linear combination of \cite{JiahuiChen:ICIP2016} (ms-cnn-lin),
end-to-end multi-scale CNN stereo of \cite{chang2018pyramid} (PSMNet), 
2-scale CNN stereo with FCN of \cite{Ichimaru:3DV2018} (ms-cnn-fcn-2),
proposed 3-scale CNN stereo with FCN (ms-cnn-fcn-3),
and transfer learned ms-cnn-fcn-3 with bubble images (ms-cnn-fcn-3 (trans)).
All CNNs were implemented with Tensorflow and trained with Middlebury 2001, 2003, 2005, 2006, 2014, and Coel Dataset,
except ms-cnn-fcn-3 (trans) was also trained with augmented dataset, and pretrained model (KITTI 2015) was used for PSMNet.
Post-processing such as SGM and LR Check were implemented with CUDA 
and achieved processing $1024\times768$ px images (maximum disparity is 256) in 30 seconds.

The target objects were placed at a distance of 50, 60, 70cm and the depth-dependent calibration was applied.
We intentionally made bubbles to interfere image capturing process.
We reproduced four bubble environments, \ie, far little bubble, far much bubble, near little bubble, and near much bubble.
In addition, no bubble scenes as reference were prepared.
We captured three pairs of images for each target with five environments.
In total, 90 images were captured.
Then, we calculated disparity map by each CNN methods, and reconstructed all the scenes and targets.
We calculated average RMSE from the GT shape of each target.

\begin{figure*}[t]
  \centering
  \centerline{\includegraphics[width=18cm]{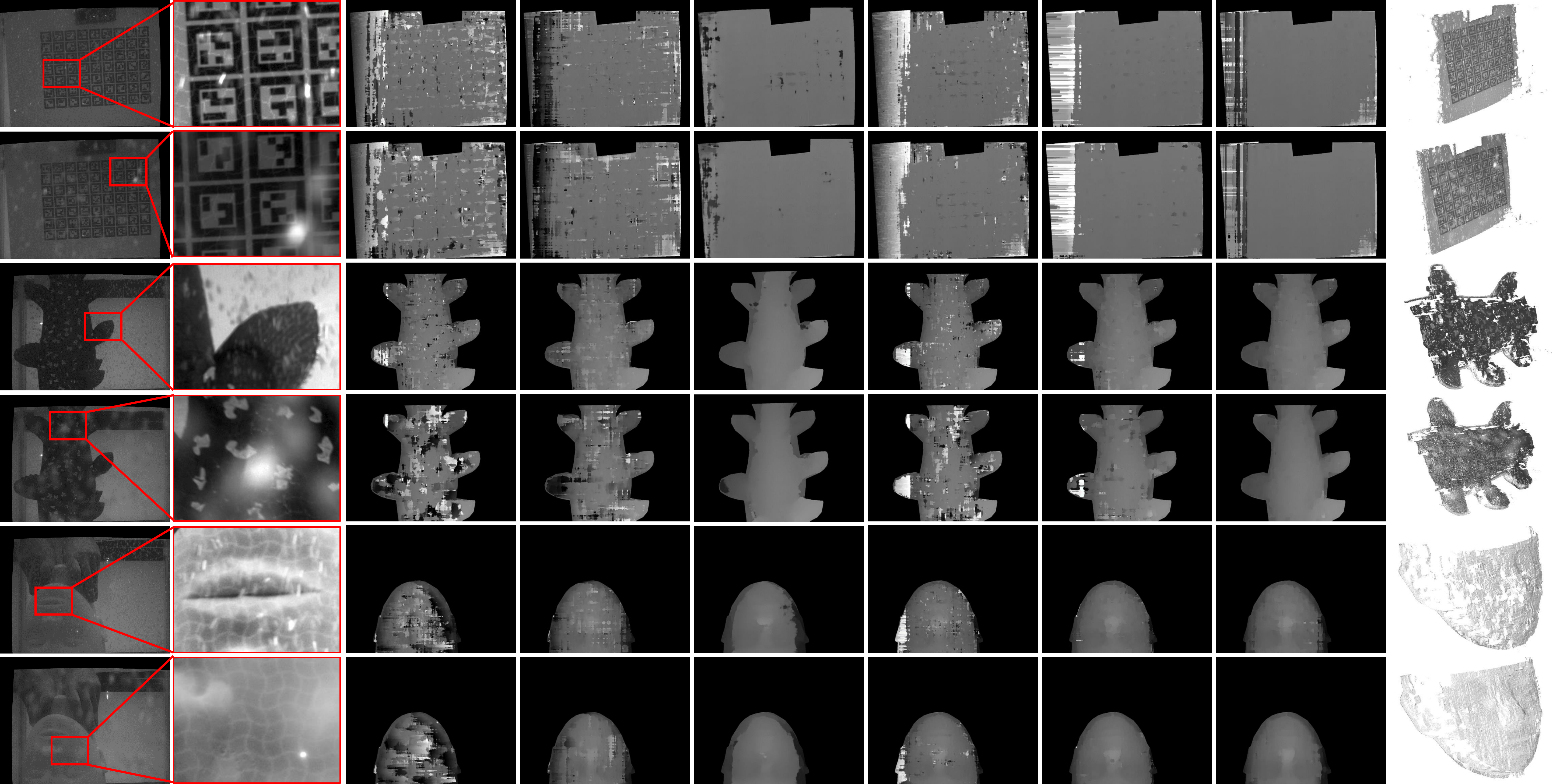}}

\caption{ Difference of disparity map between stereo methods in bubble scene. 
{\bf Left to Right: } Input image with bubbles, close-up of input, mc-cnn results~\cite{mccnn}, ms-cnn-lin results~\cite{JiahuiChen:ICIP2016}, 
PSMNet results~\cite{chang2018pyramid}, ms-cnn-fcn-2 results~\cite{Ichimaru:3DV2018}, ms-cnn-fcn-3 results, ms-cnn-fcn-3 (trans) results and reconstructed point cloud. }
\label{fig:psmnet}
\vspace{-0.0cm}
\end{figure*}

\begin{figure*}[t]
  \centering
  \centerline{\includegraphics[width=16cm]{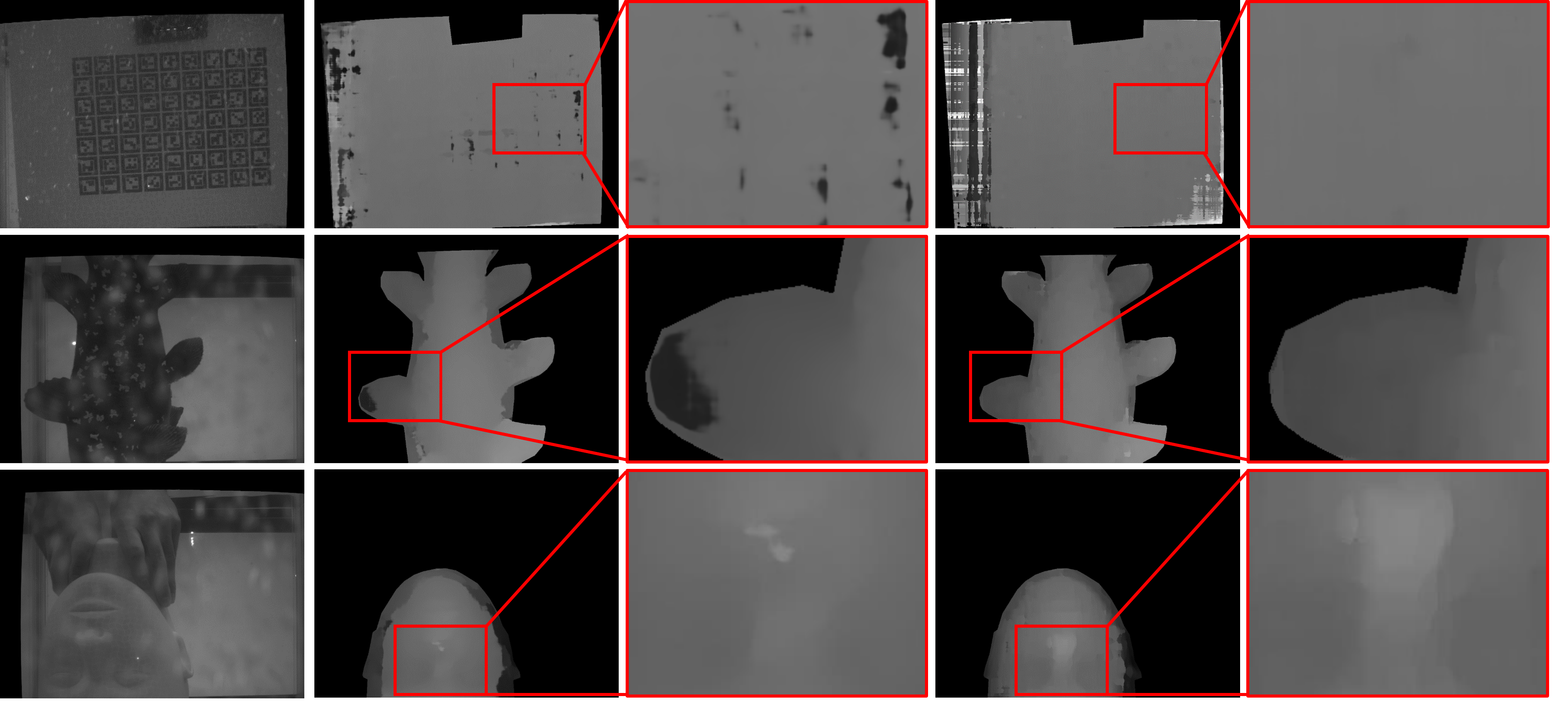}}

\caption{ Detailed comparison of disparity map between PSMNet and ours.
{\bf Left to Right: } Input image, PSMNet results, close-up of PSMNet results, ms-cnn-fcn-3 (trans) results, and close-up of ours. }
\label{fig:CNNStereo}
\vspace{-0.4cm}
\end{figure*}

\begin{figure}[t]
  \centering
  \centerline{\includegraphics[width=9.0cm]{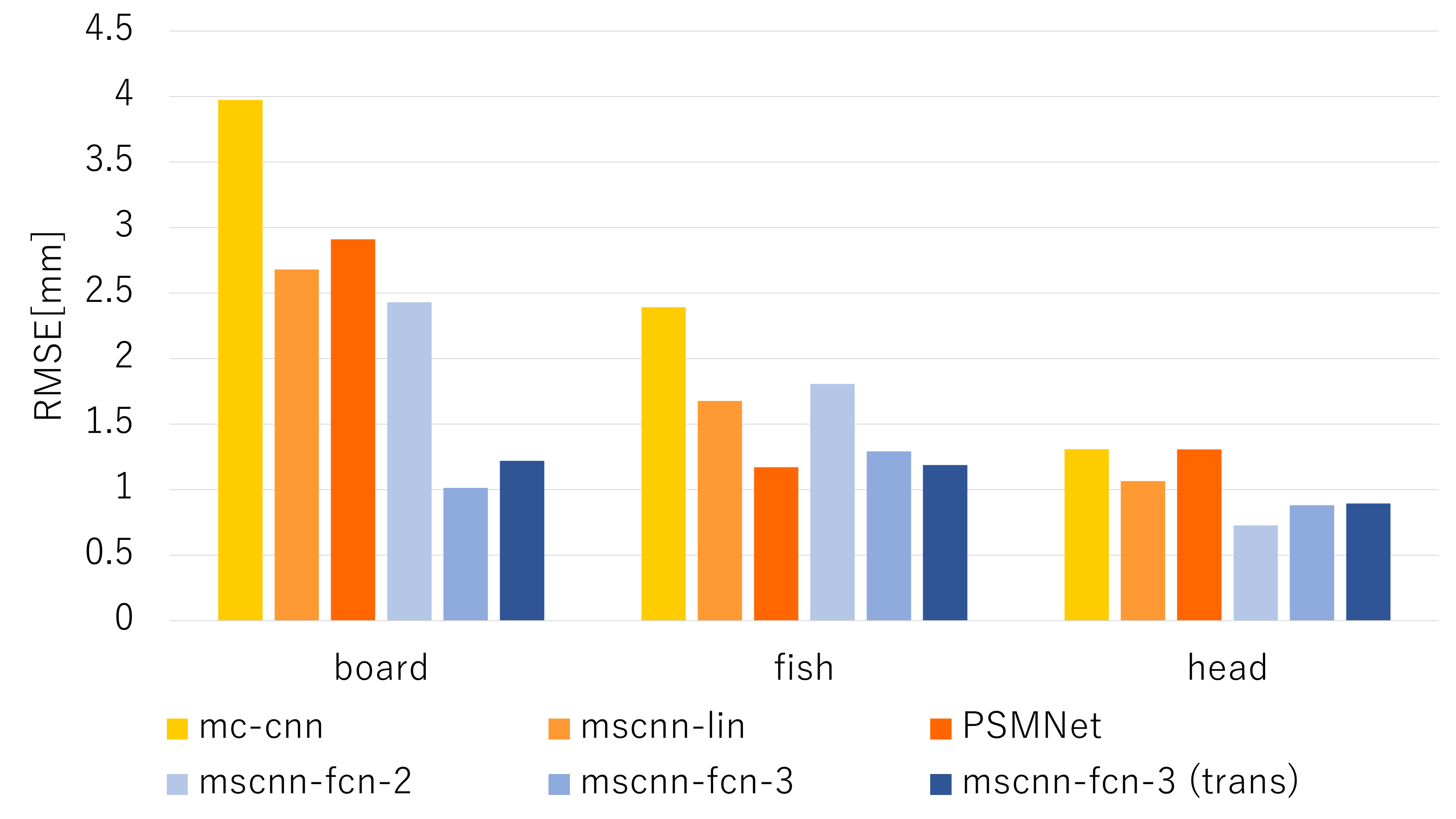}}
\caption{Comparison on proposed methods (blue bar) and previous methods(red bar). 
Proposed methods performed best in most cases. }
\label{fig:rmse}
\vspace{-0.0cm}
\end{figure}

The results are shown in Fig.~\ref{fig:rmse}.
From the graph, we can confirm that the accuracy of proposed CNN architecture is better than previous method, 
supporting the effectiveness of our method.
Fig.~\ref{fig:CNNStereo} shows examples of the reconstructed 
disparity maps for each technique confirming that shapes are recovered by 
our technique even if captured images are severely degraded by bubbles.
Further comparison between PSMNet~\cite{chang2018pyramid} and ours are shown in Fig.~\ref{fig:psmnet}.
PSMNet had difficulty to estimate correct disparity especially when the region was in occluding boundary,
whereas ours estimated correct results.

\begin{figure*}[t]
  \centerline{\includegraphics[width=18cm]{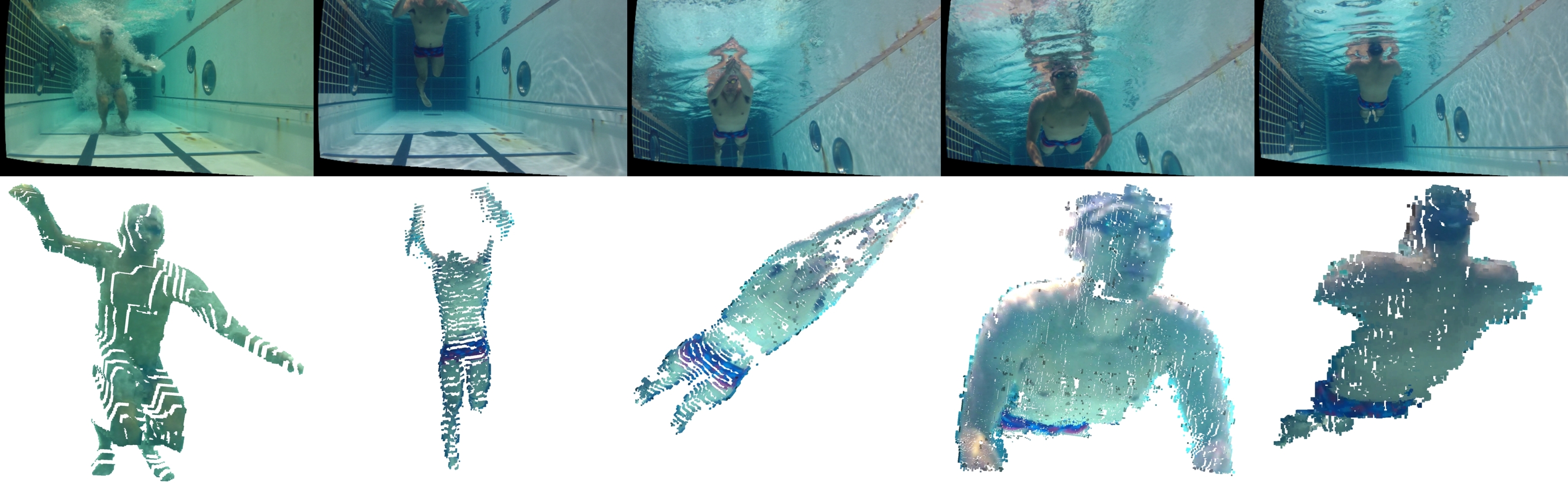}}
\caption{Swimming human experiment. Images are too small and only 
    the left-most image shows significant bubbles, but all images contain bubbles 
    in captured images.}
\label{fig:demo_human}
\vspace{-0.3cm}
\end{figure*}

\subsection{Experiments of texture recovery}
\label{ssec:texeval}

\begin{figure}[t]
  \centering
  \centerline{\includegraphics[width=9.0cm]{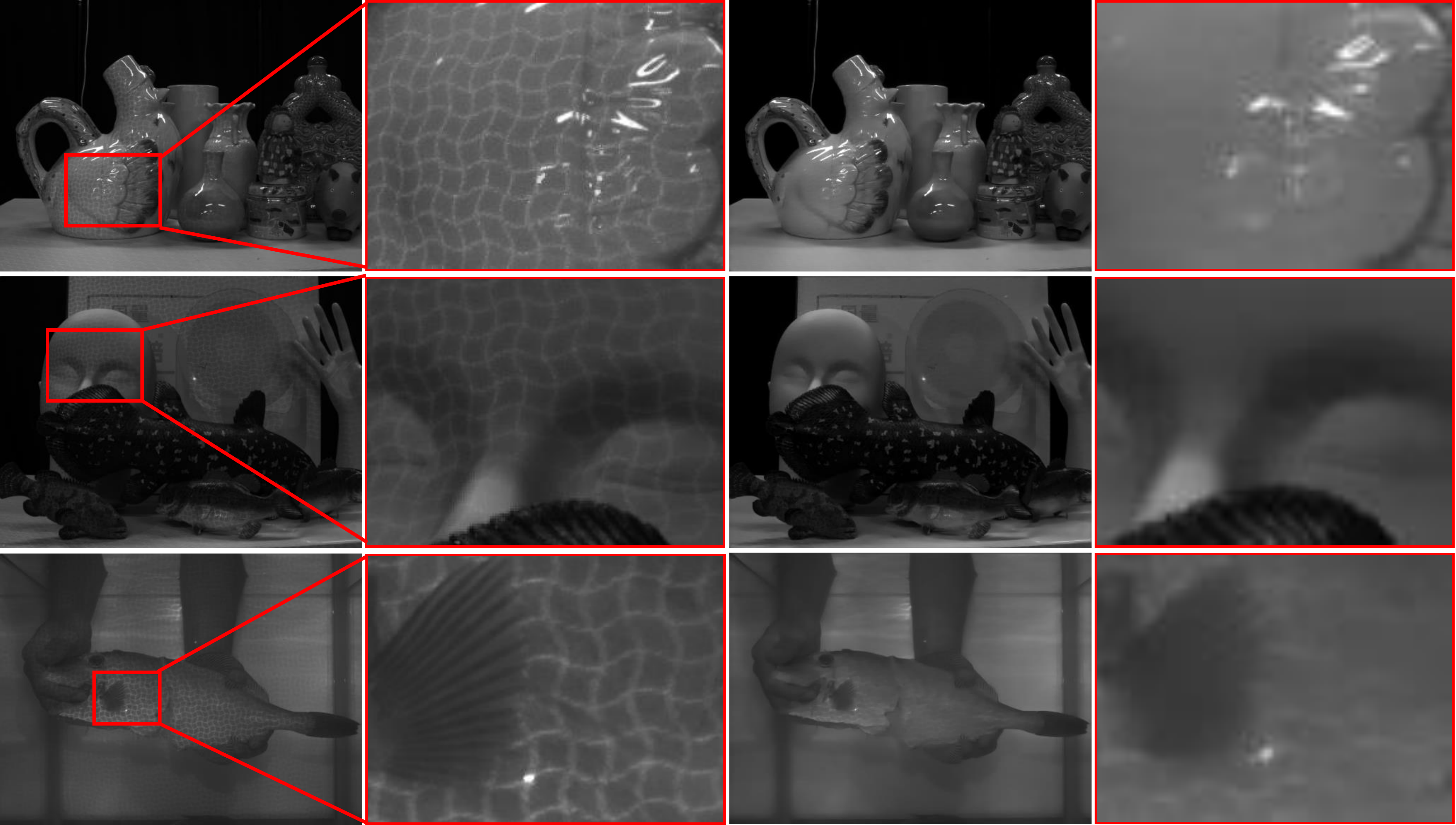}}
\caption{ Result of texture recovery experiment. 
first column is input image, 2nd column is close-up of input, 3rd column is output image, 4th column is close-up of output. }
\label{fig:texexp}
\vspace{-0.5cm}
\end{figure}

We also tested the bubble-removal and the pattern-removal techniques.
The results are shown in Fig.~\ref{fig:texexp}.
We can confirm that projected patterns are robustly removed by multi-scale CNN technique.
Top two rows are results of supervised learning, and bottom row is result of unsupervised learning,
showing unsupervised learning has enough ability to remove pattern.

\subsection{Demonstration with swimming human}
\label{ssec:demo_human}
\begin{figure}[t]
  \centerline{\includegraphics[width=6cm]{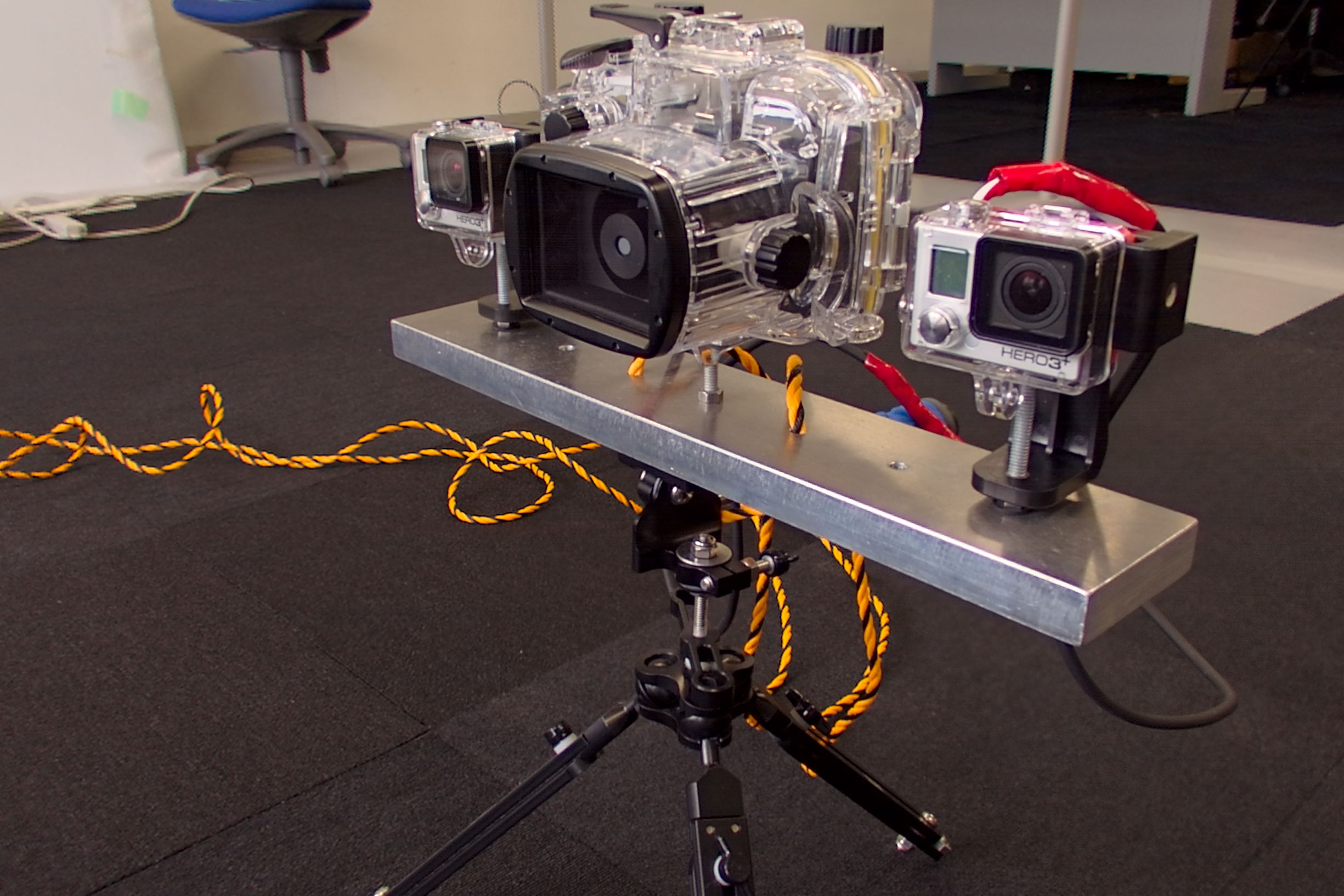}}
\caption{Experimental rig for swimming human capture.}
\label{fig:newrig}
\vspace{-0.4cm}
\end{figure}

Finally, we captured swimming human in a swimming machine, where swimmer can 
keep the same position by artificially created water flow.
We made special experimental system, which consists of low-cost commercial devices, such 
GoPro Hero 3+ for stereo camera pair with synchronous cable, and a laser pattern 
projector with battery (Fig. \ref{fig:newrig}).
We captured several swimming sequences, which is 3920 frames in 24 frames 
per seconds, \ie, 163 seconds in total.
The reconstructed results are shown in Fig.~\ref{fig:demo_human}.
In the figure, we can confirm that the 3D shape of human is successfully reconstructed by our method, 
even when the body was covered in heavy bubbles.
Since direction of optical axis of camera is almost 
parallel to human axis in this setup, reconstructed depth is largely discretized;
solution is our important future work.

\vspace{-0.0cm}
\section{Conclusion}
\label{sec:conclusion}

In this paper, we propose a robust and practical underwater dense shape 
reconstruction method using stereo cameras with a static-pattern projector.
Since underwater environments have severe conditions, such as refraction, light 
attenuation and disturbances by bubbles, we propose a CNN-based solution, such as target-object 
segmentation and robust stereo matching with a multi-scale CNN.
To acquire task-specific dataset, we created stereo dataset and special device for data augmentation which reproduces underwater environment.
We also propose a texture-recovery method using a CNN.
With our method, images with strong bubbles are robustly
recovered through comprehensive experiments showing effectiveness of our method.
Our future plan is to create underwater unmanned autonomous vehicle equipped with our system.

\section*{Acknowledgment}
This work was part supported by grant JSPS/KAKENHI 16H02849, 16KK0151, 18H04119, 
18K19824 in Japan, and MSRA CORE14.

\clearpage

{\small
\bibliographystyle{ieee}
\bibliography{STRING,JabRef,h-kawa,refs,201805303dv-r2.bib}
}

\end{document}